\documentclass[pdflatex,sn-mathphys-num]{sn-jnl}% Math and Physical Sciences Numbered Reference Style
\usepackage{graphicx}%
\usepackage{multirow}%
\usepackage{amsmath,amssymb,amsfonts}%
\usepackage{amsthm}%
\usepackage{mathrsfs}%
\usepackage[title]{appendix}%
\usepackage{xcolor}%
\usepackage{textcomp}%
\usepackage{manyfoot}%
\usepackage{booktabs}%
\usepackage{algorithm}%
\usepackage{algorithmicx}%
\usepackage{algpseudocode}%
\usepackage{listings}%
\usepackage[numbers]{natbib}
\usepackage{placeins}
\theoremstyle{thmstyleone}%
\theoremstyle{thmstyletwo}%

\theoremstyle{thmstylethree}%

\begin{document}
% --- Ajustes de espaciado fino (sin solapamiento) ---
\setlength{\textfloatsep}{8pt plus 2pt minus 2pt}   % Espacio entre figuras/tablas y texto
\setlength{\floatsep}{6pt plus 2pt minus 2pt}       % Espacio entre figuras pegadas
\setlength{\intextsep}{8pt plus 2pt minus 2pt}     % Espacio para figuras con [h]
\setlength{\dbltextfloatsep}{8pt}                   % Para figuras que ocupan dos columnas (si las hubiera)
\setlength{\abovecaptionskip}{4pt}                  % Espacio entre figura y su pie de foto
\setlength{\belowcaptionskip}{2pt}                  % Espacio después del pie de foto

\title[Parameter-Efficient Architectural Modifications for Translation-Invariant CNNs]{Parameter-Efficient Architectural Modifications for Translation-Invariant CNNs}


\author*[]{\fnm{Nuria} \sur{Alabau-Bosque}}\email{nuria.alabau@uv.es}

\author[]{\fnm{Jorge} \sur{Vila-Tomás}}%\email{iiauthor@gmail.com}

\author[]{\fnm{Paula} \sur{Daudén-Oliver}}%\email{iiiauthor@gmail.com}

\author[]{\fnm{Valero} \sur{Laparra}}%\email{iiiauthor@gmail.com}

\author[]{\fnm{Jesús} \sur{Malo}}%\email{iiiauthor@gmail.com}

\affil*[]{\orgdiv{Image Processing Lab}, \orgname{Universitat de València}, \orgaddress{\street{Carrer del Catedrátic José Beltrán Martinez}, \city{Paterna}, \postcode{46980}, \country{Spain}}}

%\affil[2]{\orgdiv{Department}, \orgname{Universitat de València}, \orgaddress{\street{Street}, \city{City}, \postcode{10587}, \state{State}, \country{Country}}}

%\affil[3]{\orgdiv{Department}, \orgname{Organization}, \orgaddress{\street{Street}, \city{City}, \postcode{610101}, \state{State}, \country{Country}}}


\abstract{Convolutional Neural Networks (CNNs) are widely assumed to be translation-invariant, yet standard architectures exhibit a startling fragility: even a single-pixel shift can drastically degrade performance due to their reliance on spatially dependent fully connected layers. In this work, we resolve this vulnerability by proposing a lightweight 'Online Architecture' strategy. By strategically inserting Global Average Pooling (GAP) layers at various network depths, we effectively decouple feature recognition from spatial location. Using VGG-16 as a primary case study, we demonstrate that this architectural modification achieves a massive 98\% reduction in trainable parameters (from 5.2M to just 82K) and a 90\% reduction in total network size ($\thicksim$
 138M to $\thicksim$ 14M). Despite this drastic pruning, our variants maintain competitive Top-1 accuracy on ImageNet (66.4\%) while doubling translational robustness, reducing average relative loss from 0.09 to 0.05. Furthermore, our analysis identifies a fundamental limit to invariance: while GAP resolves macroscopic sensitivity, discrete pooling operations introduce a residual periodic aliasing that prevents perfect pixel-level stability. Finally, we extend these findings to Perceptual Image Quality Assessment (IQA) by integrating our invariant backbones into the LPIPS framework. The resulting metric significantly outperforms the retrained baseline in generalization across the KADID-10k dataset (Spearman 0.89 vs. 0.75) and achieves a near-perfect alignment with human psychophysical response curves on the RAID dataset (Spearman 0.95). These results confirm that enforcing architectural invariance is a far more efficient and biologically plausible path to robustness than traditional data augmentation. Data and code are publicly available. The data and code are publicly available to facilitate validation and further research. \footnote{\url{https://github.com/Rietta5/Translation_Invariant_CNNs}}}




\maketitle

\section{Introduction}
\label{Intro}

Deep Convolutional Neural Networks (CNNs) have fundamentally transformed the field of computer vision, achieving superhuman performance in tasks ranging from object classification to medical image analysis~\cite{Krizhevsky12, LeCun15}. While recent years have seen a surge in attention-based architectures such as Vision Transformers (ViTs), CNNs remain profoundly relevant both in research and industrial deployment. Their inherent inductive biases—specifically local connectivity and spatial weight sharing—make them highly sample-efficient compared to ViTs, which typically lack these spatial priors and require massive datasets to generalize effectively~\cite{Dosovitskiy20}. Furthermore, modernized convolutional architectures have proven to be highly competitive with state-of-the-art transformers~\cite{Liu22}, and critically for this study, standard CNNs continue to serve as the undisputed foundational backbones for learned Perceptual Image Quality Assessment (IQA) metrics, such as LPIPS. Therefore, understanding and resolving the fundamental vulnerabilities of CNNs remains a critical open problem. A core theoretical advantage often attributed to these architectures is "translation invariance"—the ability to recognize an object regardless of its position in the image. This property is assumed to stem from the weight-sharing mechanism of convolution operations and the spatial subsampling provided by pooling layers~\cite{Goodfellow16}. However, despite this strong inductive bias, recent empirical studies have revealed a startling fragility: standard modern architectures, such as the VGG~\cite{VGG} and ResNet~\cite{ResNet} families, can exhibit drastic fluctuations in prediction confidence when inputs undergo even trivial transformations~\cite{Azulay18}. In extreme cases, a single-pixel shift can completely alter the classification output, exposing a fundamental lack of robustness in current state-of-the-art models~\cite{LPIPS}.

Conventionally, this sensitivity is mitigated through Data Augmentation~\cite{Shorten19}. By training the network on massive datasets augmented with random crops and shifts, models learn to memorize invariance~\cite{Bowers21}. While effective to a degree, this approach is akin to treating the symptom rather than the disease. Alternative structural approaches, ranging from mathematically guaranteed scattering transforms~\cite{Bruna13_scatter} to anti-aliased pooling~\cite{Zhang19}, attempt to address the signal processing roots of the problem. However, these methods often require complex retraining, specific wavelet filters, or architectural overhauls. We argue that the structural bottleneck often lies in the transition from convolutional feature maps to the fixed-size Fully Connected (FC) layers. These dense layers, which flatten the spatial dimensions, enforce a rigid dependency on absolute spatial coordinates, thereby discarding the translational equivariance built up by the convolutional stages~\cite{Lin13}.

In this work, we propose a lightweight, structural solution to this problem. Instead of relying on brute-force data augmentation or complex signal processing filtering, we introduce an "Online Architecture" approach~\cite{Bowers16}. By strategically inserting Global Average Pooling (GAP) layers into standard pretrained CNN backbones, we force the network to aggregate feature presence globally, effectively decoupling semantic content from its specific spatial location. While this architectural strategy is applicable to any CNN suffering from spatial dependency, we utilize the VGG-16 network as our primary testbed. This choice is deliberate: it provides a structurally transparent baseline and serves as the standard backbone for the LPIPS perceptual metric, allowing us to seamlessly extend our findings from classification to human psychophysics. This modification transforms the network into a translation-invariant predictor by construction, rather than by approximation

Our proposed architectural pruning allows for the removal of the massive dense layers found in standard CNN architectures. This results in a model that is not only more robust but also significantly more efficient. Specifically, it reduces the trainable parameter count by over 98\% (from approximately 5.2M in our baseline to just 82K), while simultaneously decreasing the total network parameters from the $\sim$ 138M of a standard VGG-16 to roughly $\sim$ 14M.

Furthermore, the implications of this work extend beyond classification into the domain of Image Quality Assessment (IQA). While traditional metrics like SSIM~\cite{SSIM} or biologically inspired models based on information theory~\cite{Malo97} and divisive normalization~\cite{Laparra16_NLPD} attempt to explicitly model the Human Visual System (HVS), learned metrics like LPIPS~\cite{LPIPS} have become the standard for evaluating generative models. However, standard LPIPS inherits the translational sensitivity of its backbone. This creates a disconnect: imperceptible shifts are penalized as severe errors, a failure often masked in standard IQA benchmarks but critical in psychophysical terms~\cite{Martinez19}. By integrating our invariant backbones into the LPIPS framework, we aim to align the metric with biological plausibility.

Our contributions are summarized as follows: (1)~We propose and evaluate GAP-modified CNN variants that achieve superior translation robustness on ImageNet. Specifically, our approach effectively mitigates the performance drop observed in standard models by reducing the average relative loss from 0.09 to 0.05 under spatial shifts, without requiring extensive retraining. (2)~We demonstrate that replacing dense layers with GAP results in a 98\% reduction in trainable parameters, proving that robustness does not require increased model complexity. (3)~We successfully apply the invariant backbone to the LPIPS metric. We validate this new metric against human data, demonstrating a significant leap in generalization across unseen distortions, where our variant improves the Spearman correlation from 0.75 (standard retrained LPIPS) to 0.89. (4)~We replicate the human suprathreshold response curves measured on the RAID dataset~\cite{RAID}, where the modified model reproduces the human curves more accurately. This confirms that our invariant architecture strictly aligns with the biological reality of the Human Visual System. Consequently, it provides a much more reliable and perceptually accurate metric for downstream tasks (such as evaluating generative models or compression algorithms), as it correctly mirrors human tolerance to trivial spatial shifts; and (5)~We provide a rigorous analysis of the residual sensitivity, identifying the discrete sampling of pooling layers—and the resulting aliasing—as the fundamental limit to achieving perfect pixel-level invariance in CNNs.

The remainder of this article is organized to guide the reader from theoretical foundations to perceptual applications. Section~\ref{sec:Previo} establishes the necessary background on geometric stability, distinguishing between equivariance and invariance. Section~\ref{sec:Mejora} introduces our core architectural proposal, detailing the strategic GAP insertion schemes and their parametric efficiency. The experimental framework for classification is outlined in Section~\ref{sec:Experimental}, paving the way for Section~\ref{sec: results_clas}, where we analyze the models' performance. Here, we first isolate the intrinsic aliasing problem of discrete pooling using the MNIST dataset, and subsequently evaluate large-scale robustness on ImageNet. Section~\ref{sec:results_IQA} extends these findings beyond classification into Image Quality Assessment (IQA), embedding our invariant backbones into the LPIPS metric and validating the perceptual alignment against human psychophysical data. Finally, Section~\ref{sec:Discussion} contextualizes our findings within the broader field, and Section~\ref{sec:conclusion} draws the final conclusions.

\section{Previous Concepts}
\label{sec:Previo}

Before addressing the architectural limitations of standard CNNs, it is necessary to establish a precise theoretical framework regarding geometric stability. In this section, we first formalize the critical distinction between invariance and equivariance. Understanding this dichotomy is essential to grasp why standard subsampling operations (pooling)—which are equivariant by nature—fail to guarantee the translational invariance required for robust classification. Then, we define affine transformations, which constitute the mathematical basis for the geometric perturbations used throughout this work to stress-test model stability.

\subsection{Invariant and equivariant transformations}

Since this paper focuses solely on images that have been modified by translations, we first define the concepts of equivariant and invariant transformations. In this context, given an image $A$ and the group of translational transformations $G$, a function or system $f$ is said to be \textbf{equivariant to the transformation $g \in G$} if the output of the function $f$ maintains that transformation $g$~\cite{lecun1995convolutional, NIPS2014_f9be311e, fukusima}. That is:

\begin{equation}
\begin{aligned}
    B & = f(A) \\
    g(B) & = f(g(A))
  \end{aligned}
\end{equation}

On the other hand, a function or system $f$ is said to be \textbf{invariant to the transformation} if the output of the function $f$ is not modified~\cite{Azulay18, Gong14, Eric18}. That is:

\begin{equation}
\begin{aligned}
    B & = f(A) \\
    B & = f(g(A))
  \end{aligned}
\end{equation}

\begin{figure}[!h]
	\centering
	\includegraphics[width=0.7\linewidth]{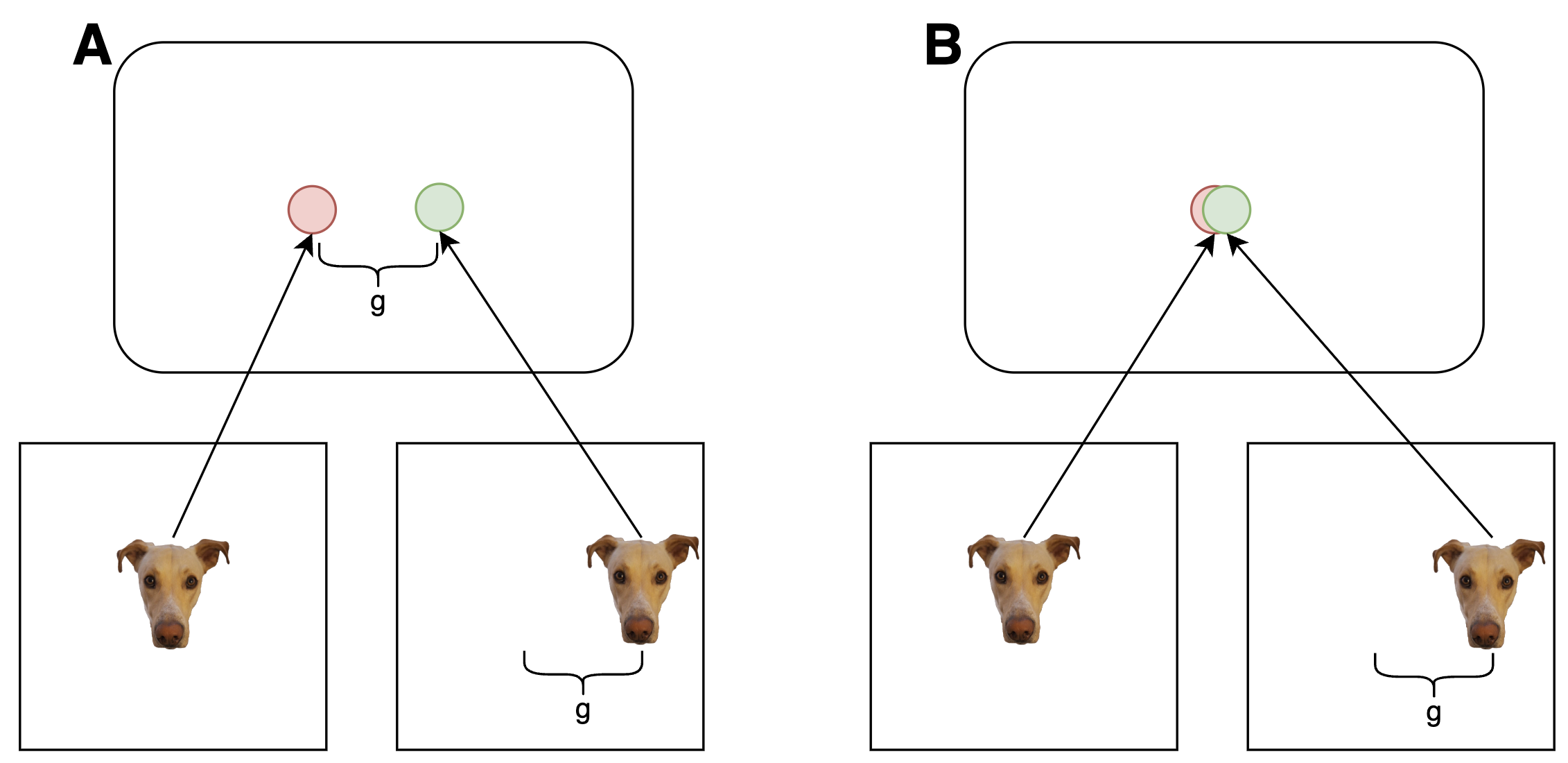}
	\caption{Diagram of the models with equivariant (A) and invariant features (B). In A, when the input image is translated, the same translation is maintained in the internal domain. In B, when the input image is translated, the representation in the internal domain is unaffected.}
	\label{fig:EquiInv}
\end{figure}

A diagram of both types of models is shown in Figure~\ref{fig:EquiInv}. In an equivariant model, a transformation in the image domain leads to an equally transformed point in the problem domain, while an invariant model produces the same output for both the transformed and untransformed inputs.

\subsection{Affine Transformations}
\label{subsec:trasnformations}

An affine transformation or affine application (also called affinity) between two affine spaces is a transformation that satisfies Equation~\ref{eq:affine}.

\begin{equation}
    F: v \to Mv + b
    \label{eq:affine}
\end{equation}

Where $v$ can be any vector and the affine transformation is represented by a matrix $\mathbf{M}$ and a vector $\mathbf{b}$ satisfying the following properties: first, it maintains the collinearity (and coplanarity) relations between points and, second, it maintains the ratios between distances along a line. (Equation~\ref{eq:affine1})\\

\begin{equation}
    i'(x) = i(Mx + b)
    \label{eq:affine1}
\end{equation}

Some examples of affine transformations are geometric contraction, expansion, dilation, reflection, rotation, or shear.

\section{Proposed improvements}
\label{sec:Mejora}

This section details the specific architectural modifications proposed to enhance the translation invariance capacity of pretrained classification networks. Our proposal is a general "Online Architecture" approach that leverages the spatial averaging properties of the Global Average Pooling (GAP) layer. To empirically validate this methodology, we selected the VGG-16 network as our fundamental backbone. Beyond its architectural simplicity, which allows for clear isolation of the pooling aliasing effects, VGG-16 is the foundational feature extractor for the standard LPIPS metric. Demonstrating our method on VGG-16 ensures direct applicability to state-of-the-art Image Quality Assessment (Section~\ref{sec:results_IQA}).

\subsection{Invariant Modification Schemes}

The fundamental problem with traditional CNN classification is the reliance of the final fully connected (FC) layers on the precise spatial location of features, which is the root cause of translational sensitivity. The GAP operation, which averages activations spatially across an entire feature map, effectively decouples feature existence from its exact position.

%Nomenclature: For the sake of brevity and clarity throughout the rest of this manuscript, we adopt the following simplified nomenclature to refer to the studied architectures: the baseline VGG-16 is denoted as Base; the variant with GAP layers after each pooling block as Multi; the variant with a single GAP replacing the final dense layers as Final; and the flattened multi-stage variant as Flat.

\begin{figure}[!h]
	\centering
	\includegraphics[width=0.8\linewidth]{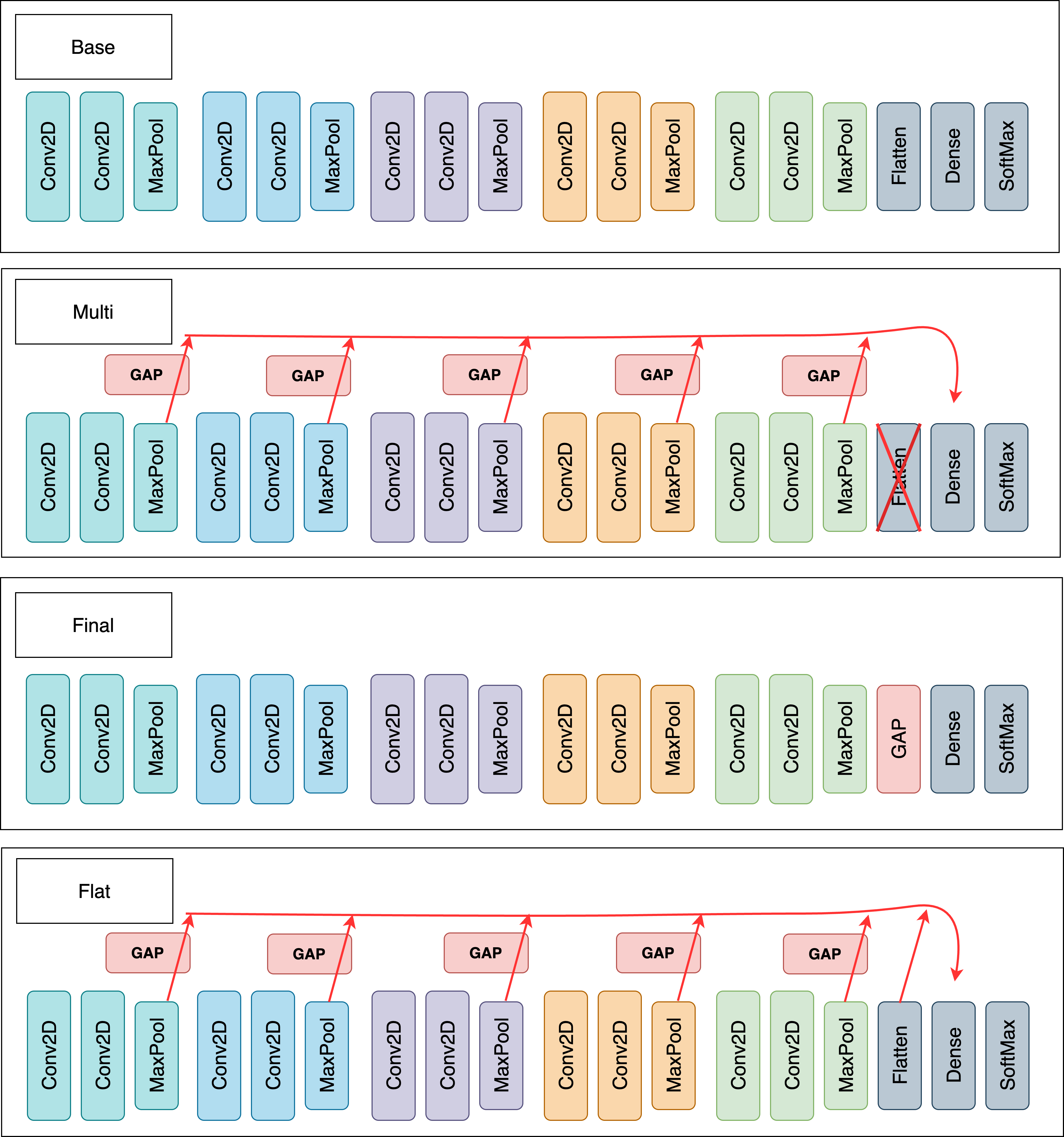}
	\caption{Proposed Architectural Modifications for Enhanced Translation Invariance in VGG-16. Diagram illustrating the baseline VGG-16 architecture and the four proposed schemes for introducing translation-invariant paths. The modifications involve inserting Global Average Pooling (GAP) layers at various depths (after selected convolutional blocks). The resulting feature vectors from these GAP layers bypass the remainder of the fully convolutional layers and are then concatenated at the final classification layer, creating parallel invariant routes that enhance robustness while minimizing additional parameters.}
	\label{fig:mejoras}
    %\vspace{-0.4cm}
\end{figure}

Our core modification involves strategically inserting GAP layers within the Base architecture to generate translation-invariant feature vectors. As visually detailed in Figure~\ref{fig:mejoras}, we developed and tested the following schemes:

\begin{itemize}
\item \textbf{Base:} The original VGG-16 architecture used for benchmarking.

\item \textbf{Multi:} In this variant, multiple GAP layers are inserted after each MaxPool operation. The resulting invariant feature vectors bypass the remaining standard convolutional layers and are concatenated before the final classification layer.

\item \textbf{Final:} This model simplifies the structure by replacing the entire block of final dense layers of the Base model with a single GAP layer followed by a single Dense layer.

\item \textbf{Flat:} This variant inserts the GAP layers similarly to the Multi model, but utilizes a Flatten layer before concatenation. This results in a higher parameter count but preserves specific structural information from the original VGG feature maps.
\end{itemize}

By concatenating these invariant summaries (in the Multi and Flat cases) or replacing the dense head entirely (in the Final case), the network is provided with robust, position-independent information that anchors the classification decision.

\subsection{Parameter Efficiency}

A key advantage of our proposal is the parametric efficiency achieved by replacing standard fully-connected layers with the GAP operation, which adds few or no trainable parameters. Table~\ref{tab:Parametros} provides a detailed breakdown of the parameter counts for the Base model and the three modified variants.

\begin{table}[h]
\caption{A breakdown of the total, trainable, and non-trainable parameters for the Base network and the three GAP-modified variants: Multi, Final, and Flat. The Multi and Final variants achieve significant architectural modification and enhanced translation invariance while drastically reducing the number of trainable parameters compared to the Base and Flat.}
\label{tab:Parametros}
\begin{tabular}{lccc}
\toprule
Model & Total parameters & Trainable parameters & Trainable params w.r.t. baseline \\ 
\midrule
Base & 19,957,728 & 5,243,040 & $100\%$ \\ 
Multi & 14,950,368 & 235,680 & $4\%$ \\ 
Final & 14,796,768 & 82,080 & $1.5\%$ \\ 
Flat & 20,193,248 & 5,478,560 & $104\%$ \\ 
\bottomrule
\end{tabular}
\vspace{-0.4cm}
\end{table}

As shown in Table~\ref{tab:Parametros}, the Multi and Final variants achieve a drastic reduction in the number of trainable parameters compared to the Base model. Specifically, the Final model reduces the number of trainable parameters to only 82,080 from the baseline's 5,243,040, a reduction of over 98\%. This remarkable efficiency demonstrates that the proposed architectural changes not only enhance translation invariance but also result in significantly lighter models, addressing the issue of high computational resource requirements. The empirical results in Section 4 confirm that this parameter reduction in the Final and Multi models does not compromise performance but rather leads to superior stability.

\section{Experimental Setting}
\label{sec:Experimental}

In this section, we review the selected databases and affine transformations that will be used in the experiments. The code used is made available publicly to allow researchers and practitioners to test their own metrics. \url{https://github.com/Rietta5/Translation_Invariant_CNNs}

\subsection{Datasets}
\label{subsec:datasets}
%\hhh{Subsubheading}

The empirical evaluation in this study relies on several distinct datasets, each serving a specific analytical purpose across our classification and quality assessment experiments:

\begin{itemize}
    \item MNIST: As a controlled, low-resolution environment, we utilize the classic MNIST dataset. Rather than for benchmark performance, MNIST is specifically employed in our initial experiments to isolate and visually demonstrate the sub-pixel aliasing effects caused by discrete pooling layers, completely free from the textural complexities of natural images.
    
    \item ImageNet Subset: To evaluate the proposed architectures in a complex, real-world domain while maintaining computational tractability, we curated a balanced, representative subset of the ImageNet-1K dataset for our main robustness benchmarks. Specifically, we selected 160 uniformly represented classes, comprising 60,000 training images and 10,000 test images.

    \item IQA and Psychophysical Datasets: For the perceptual evaluation (detailed in Section 6), we employ standard Image Quality Assessment databases (TID08~\cite{TID08}, TID13 ~\cite{TID13}, and KADID-10k~\cite{kadid}) to test generalization, alongside the RAID dataset~\cite{RAID} to validate our models against human psychophysical response curves under geometric distortions.
\end{itemize}

\subsection{Image Transformations}

The core hypothesis of this work centers exclusively on translation invariance. Therefore, the primary geometric perturbation applied across our experiments is spatial displacement. However, to rigorously validate the specificity of our architectural modifications—ensuring that the GAP mechanism isolates translation rather than indiscriminately altering the network's general geometric sensitivity—we also conducted control experiments using rotation and scaling. The detailed analysis of these non-translational transformations is relegated to Appendix~\ref{app:apndB}.

\begin{center}
\begin{figure}[!h]
\centering
			\includegraphics[width=0.58\linewidth]{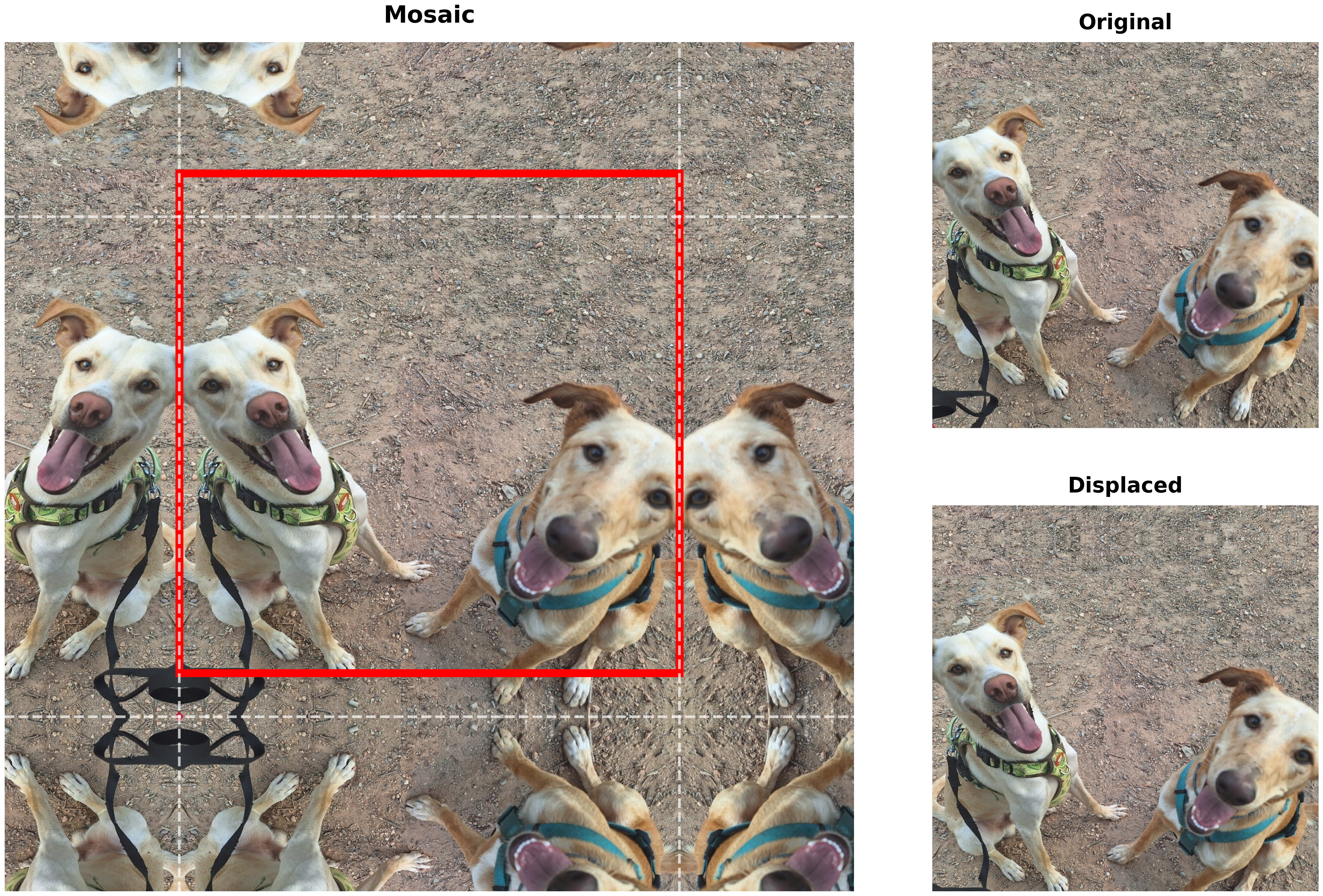}	
		
		\caption{Translation transformation and mosaic padding technique to evaluate spatial invariance without introducing zero-padding artifacts. (Left) The original image is tiled into a larger extended canvas. The required spatial displacement is applied to this entire canvas, and a center patch matching the original dimensions (indicated by the red bounding box) is extracted. (Right) Example of the final result, comparing the original image with its translated counterpart, completely free of artificial boundary cues.}
		\label{fig:distorsiones_Traslation}
        %\vspace{-0.4cm}
	\end{figure}
\end{center}

When applying these affine transformations to the ImageNet subset, a common issue is the introduction of artificial black borders (zero-padding artifacts) or the loss of the central semantic element. To mitigate this, we implemented a mosaic-padding technique: the original image is first tiled into a larger mosaic, the geometric transformation is applied to this extended canvas, and finally, a center patch matching the original dimensions is extracted. This guarantees artifact-free boundaries, as illustrated in Figure~\ref{fig:distorsiones_Traslation}.

\begin{center}
\begin{figure}[!h]
\centering
			\includegraphics[width=1\linewidth]{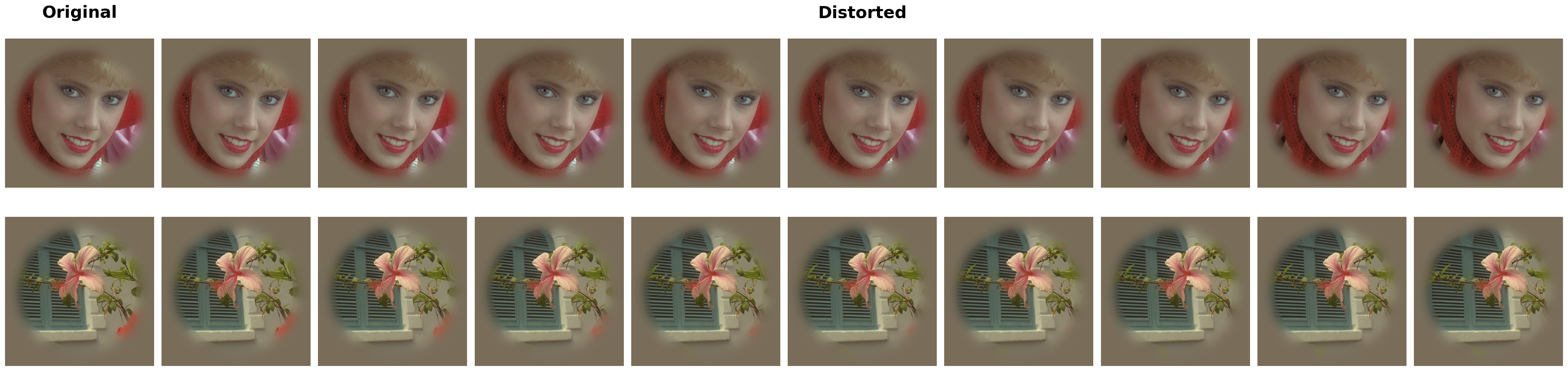}	
		
		\caption{Stimuli examples from the RAID dataset used for the psychophysical and IQA validation. To prevent artificial edge boundary cues from influencing the perceptual assessment of geometric changes, the images are masked with a circular aperture (soft windowing). The figure displays two different stimuli (rows), starting with the pristine Original image (left) followed by a sequence of increasingly Translated counterparts across the distortion range.}
		\label{fig:distorsiones_RAID}
        \vspace{-0.4cm}
	\end{figure}
\end{center}

The specific transformation parameters utilized in this study are defined as follows:

\begin{itemize}
    \item Translation (Primary Focus): Displacements on the vertical and horizontal axes (and their combinations) with an amplitude of 10, 25, or 50 pixels depending on the image size in MNIST experiments and 50 pixels in ImageNet experiments (equivalent to $0.3^{\circ}$ of translation in psychophysical terms) in each direction.
    \item Rotation and Scale (Control Experiments): Rotations ranging from $0^{\circ}$ to $20^{\circ}$, and scale factors from 0.1 to 2.0. These are used strictly to test the boundary conditions of our invariant architectures.
    
    Additionally, Figure~\ref{fig:distorsiones_RAID} presents the stimuli in the RAID dataset for the psychophysical validation. Note the use of a circular aperture in the RAID examples; this windowing technique is essential to eliminate boundary cues, ensuring that the perceptual assessment focuses exclusively on the internal geometric distortion rather than edge artifacts.
    
\end{itemize}
\vspace{-0.4cm}

\vspace{-0.4cm}

\section{Results in classification problems}
\label{sec: results_clas}

In this section, we empirically evaluate the translation invariance capabilities of the proposed architectures. To provide a clear and intuitive understanding of the underlying mechanics, we first present a controlled experiment using the MNIST dataset. This initial analysis visually isolates the fundamental aliasing problem inherent to discrete pooling operations, demonstrating why standard CNNs struggle with sub-pixel invariance. Following this theoretical grounding, we systematically benchmark our modified networks (Multi, Final, and Flat) against the Base architecture on the large-scale ImageNet dataset. This allows us to quantify both the robustness gained under spatial shifts and the efficiency of the models in a highly complex, real-world scenario

\subsection{Visualizing Translation Sensitivity and Pooling Aliasing}
\label{sec: results_MNIST}

 While the implementation of Global Average Pooling (GAP) dramatically enhances the translation invariance of VGG-based models, residual periodic patterns of sensitivity remain, particularly at higher input resolutions. This section demonstrates that these patterns are not noise but are directly traceable to the interaction between the GAP layer and the preceding discrete pooling operations.
 
\begin{figure}[!h]
	\centering
	\includegraphics[width=0.75\linewidth]{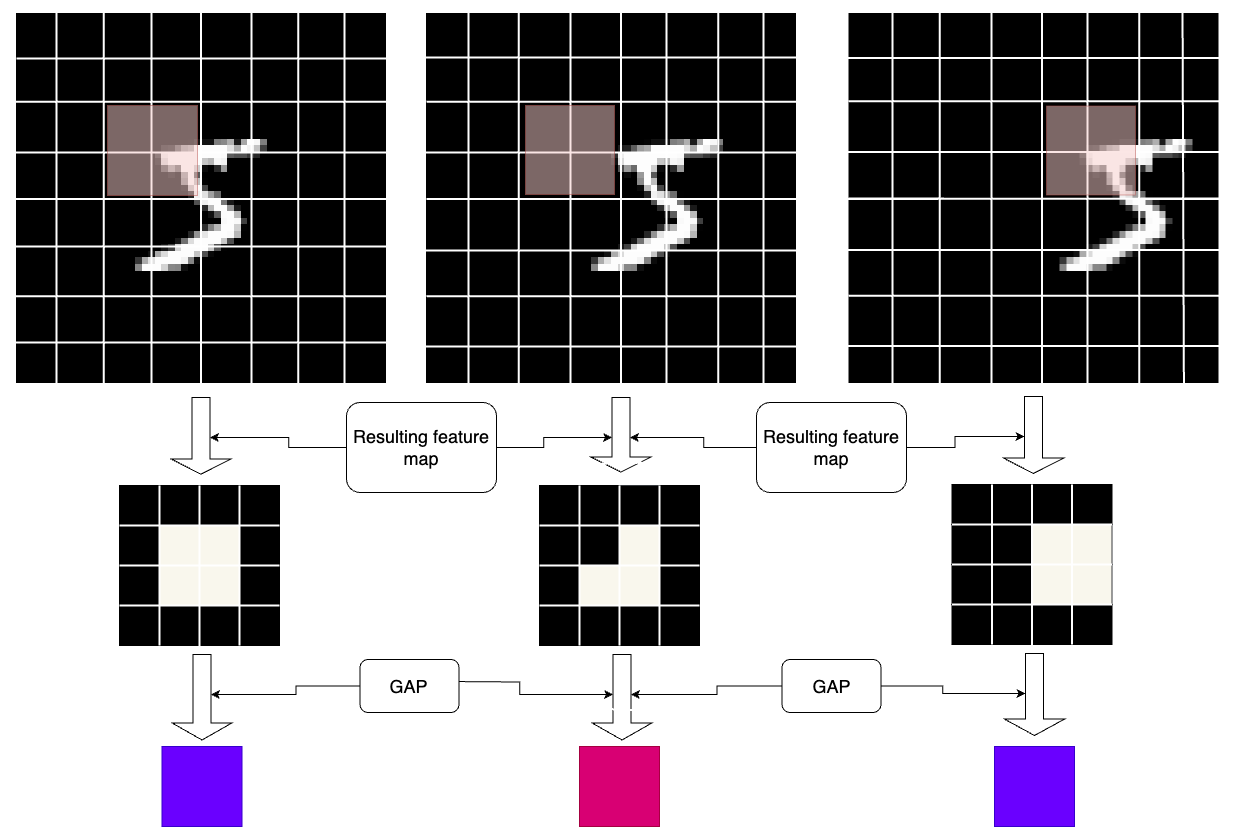}
	\caption{Diagram of the behavior of the pooling layer combined with the GAP layer. Initially, the centered image generates a feature map and then a summary value after the GAP. When the original image is shifted by one pixel, the pooling windows sample different spatial regions, generating an altered feature map and, consequently, a new GAP summary value. Finally, when we shift two pixels, the pooling falls in the same positions as in the centred image and the GAP summarises in the same way as at the beginning.}
	\label{fig:Diagrama}
    %\vspace{-0.4cm}
\end{figure}

 The hypothesized mechanism for this phenomenon is illustrated conceptually in Figure~\ref{fig:Diagrama}. In an ideal invariant network, a shift in the input image should not alter the feature map summary value generated by GAP. However, the pooling layer acts as a spatial filter with fixed windows. When the input image is shifted by a non-congruent distance $s$ (e.g., $s=1$), the pooling windows sample entirely new regions of the feature map, yielding a significantly different output and consequently a distinct summary value from GAP, causing an accuracy drop. Conversely, when the shift $s$ is an exact multiple of the pooling kernel size $k$ (e.g., $s=2$), the pooling windows realign with the original sampling grid, generating a feature map similar to the original one and recovering the classification accuracy.
 
\begin{figure}[htbp]
    \centering
    \includegraphics[width=0.75\linewidth]{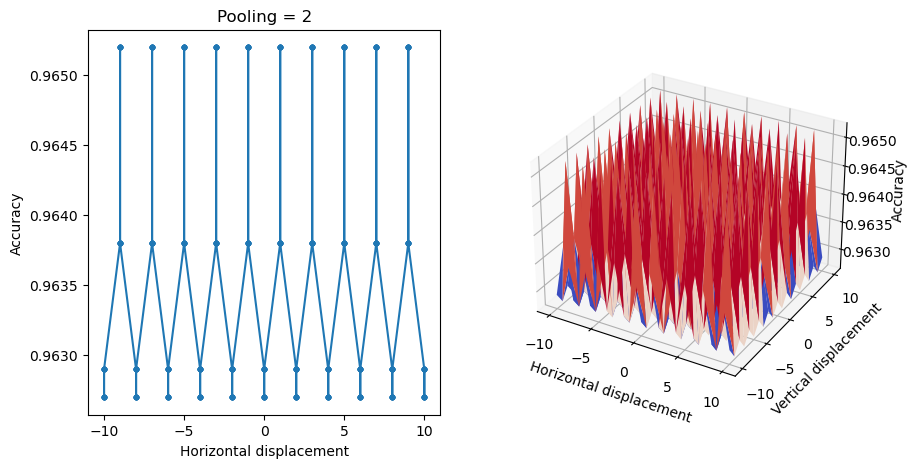}\\[1ex]
    \includegraphics[width=0.75\linewidth]{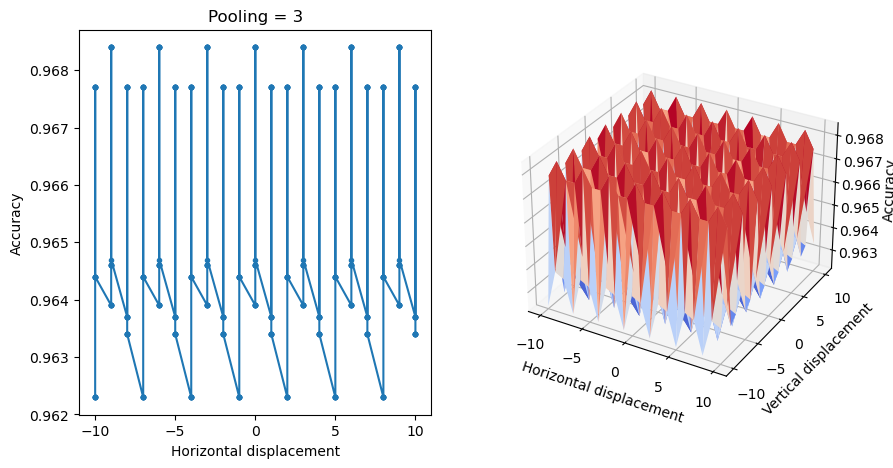}\\[1ex]
    \includegraphics[width=0.75\linewidth]{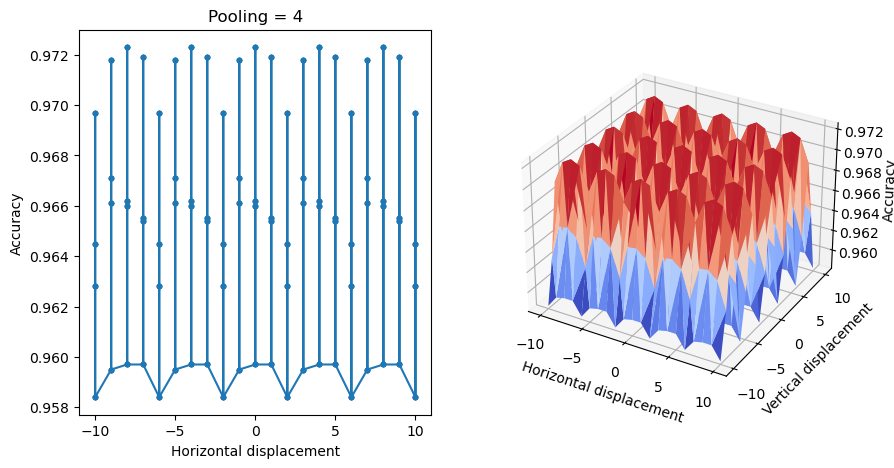}
    \caption{Accuracy Oscillations Induced by Pooling Size and Translational Shift (Toy Problem). Two-dimensional (left column) and three-dimensional (right column) surface plots illustrating the classification accuracy of a simple model equipped with a GAP layer, tested under horizontal and bidirectional translational displacements on the MNIST dataset. The experiment is repeated for different pooling kernel sizes: $\text{Pooling}=2$ (Top row), $\text{Pooling}=3$ (Middle row), and $\text{Pooling}=4$ (Bottom row). The plots demonstrate a periodic pattern of accuracy loss and recovery. Specifically, accuracy reaches a local maximum when the shift is an integer multiple of the pooling kernel size $k$ (e.g., $2, 4, 6, \dots$ pixels for $\text{Pooling}=2$), and drops to a local minimum when the shift causes maximal misalignment. This behavior confirms that the interaction between the discrete nature of the pooling operation and the spatial averaging of the GAP layer is the source of the observed translational sensitivity.}
    \label{fig:toy_models}
    \vspace{-0.4cm}
\end{figure}

 This periodic accuracy oscillation was experimentally validated using a controlled toy problem on the MNIST dataset, the results of which are visualized in Figure~\ref{fig:toy_models}. The 2D and 3D plots explicitly confirm that classification accuracy follows a periodic function of the translational displacement, with the period directly corresponding to the size of the pooling kernel $k$. The accuracy peaks occur precisely when the shift aligns with the pooling stride, and the valleys occur at misalignment points.
 
\begin{figure}[!h]
	\centering
	\includegraphics[width=0.75\linewidth]{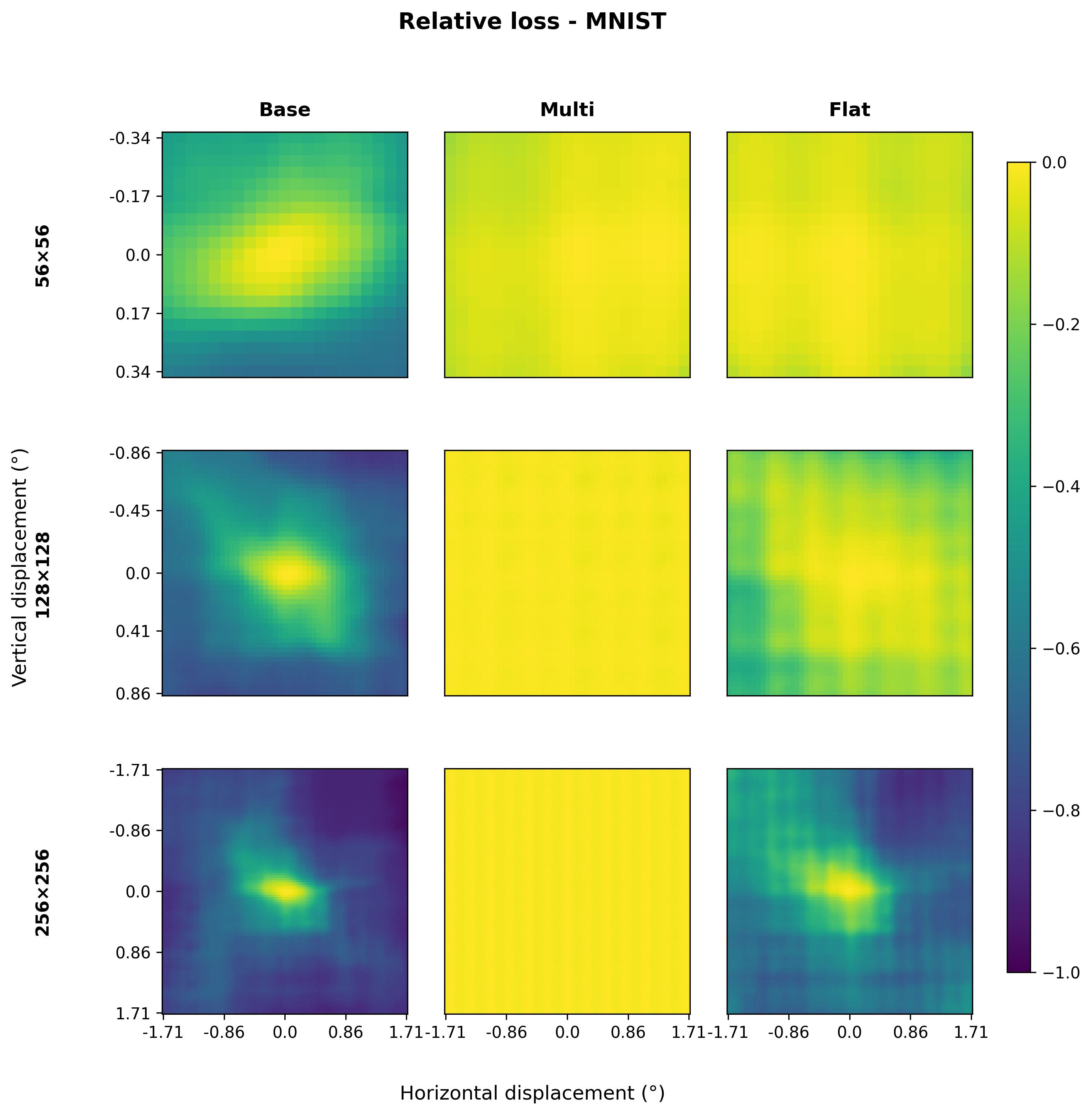}
	\caption{Heatmaps illustrating the relative loss in Top-1 classification accuracy compared to the optimally centered image (zero displacement) for the baseline VGG16 model and two prominent GAP-modified variants, Multi and Flat. The results are presented for three different input image resolutions (rows: $56 \times 56$, $128 \times 128$, and $256 \times 256$). The loss is normalized across all resolutions and models for comparative visualization. The Base model exhibits high, concentrated relative loss that increases with resolution. The VGG16GAP model achieves near-perfect translation invariance across all resolutions (flat yellow heatmaps), while Multi shows periodic vertical patterns, suggesting residual translational sensitivity related to the pooling structure.}
	\label{fig:Collage_N}
    %\vspace{-0.4cm}
\end{figure}

\begin{figure}[!h]
	\centering
	\includegraphics[width=0.75\linewidth]{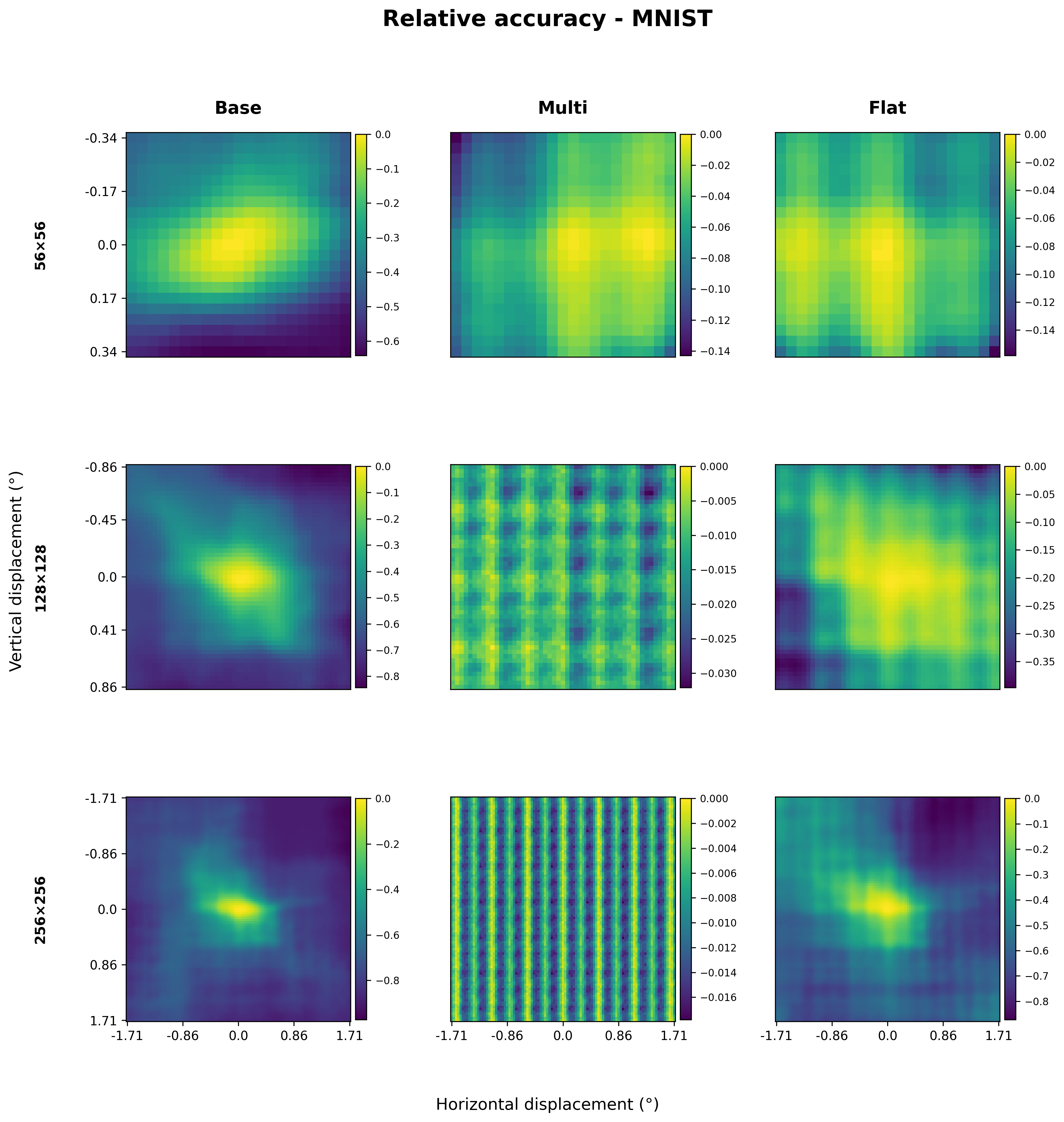}
	\caption{Heatmaps illustrating the absolute relative loss in Top-1 classification accuracy for the Base model, Multi, and Flat, tested across three different input resolutions ($56 \times 56$, $128 \times 128$, and $256 \times 256$). Crucially, the color bars are independently scaled to reflect the raw magnitude of the relative loss ($\text{Accuracy}_{0,0} - \text{Accuracy}_{x,y}$) for each model. The baseline shows the largest magnitude of loss (up to -1.0) concentrated at the center. The Multi model achieves a remarkable level of invariance, with absolute loss magnitudes two orders smaller than the baseline (e.g., maximum loss of only $\approx -0.04$ at $128 \times 128$). Most notably, at higher resolutions ($128 \times 128$ and $256 \times 256$), the Multi model exhibits clear, periodic vertical patterns across the displacement space, demonstrating residual translational sensitivity tied to the underlying convolutional/pooling structure, despite the overall high level of invariance achieved.}
	\label{fig:Collage_SN}
    %\vspace{-0.4cm}
\end{figure}

 The consequences of this microscopic architectural behavior are then magnified and observed in the full VGG-16 models. Figure~\ref{fig:Collage_SN} provides the most compelling visual evidence. While the overall loss magnitude for the Multi variant is drastically reduced compared to the baseline, the heatmaps for $128 \times 128$ and $256 \times 256$ resolutions reveal distinct, persistent vertical bands of higher relative loss. These bands represent the macroscopic manifestation of the pooling misalignment observed in Figure~\ref{fig:toy_models}. The vertical alignment of the bands suggests that the horizontal displacement is particularly effective at disrupting the pooling window boundaries, leading to the localized spikes in prediction failure. Although Figure~\ref{fig:Collage_N} (Normalized Relative Loss) shows that the Multi variant is visually flat across the entire space, the un-normalized visualization in Figure~\ref{fig:Collage_SN} is essential, as it isolates and identifies the subtle architectural weakness that prevents the model from achieving perfect, non-periodic translation invariance.

\subsection{Large-Scale Robustness Evaluation on ImageNet}
\label{sec: results_Imagenet}

The robustness of the baseline model and the proposed variants was systematically evaluated on the ImageNet validation set by subjecting the images to a wide range of translational displacements. The results are analyzed through numerical metrics and visualizations in two-dimensional heatmaps.

Table~\ref{tab:mean_std} summarizes the overall classification performance (Mean Accuracy) and translational robustness ("Loss w.r.t. center") averaged across the entire displacement space. The baseline VGG16 shows a Mean Accuracy of $0.64$ and a relatively high standard deviation of $0.04$. In contrast, the Multi and Final models not only maintain comparable or superior Accuracy ($0.636$ and $0.664$, respectively), but also demonstrate a substantial improvement in the robustness metric. The Multi variant reduces the average loss (Loss w.r.t. center) to $0.05$ with a standard deviation of $0.02$, representing superior stability and less performance degradation under translation compared to the baseline (loss of $0.09$ and standard deviation of $0.05$). This numerical analysis establishes a clear advantage for the GAP variants in terms of consistency and reduced performance drop against translation.

\begin{table}[htbp]
\caption{Summary of ImageNet Classification Performance and Translational Robustness. Mean $(\mu)$ and standard deviation $(\sigma)$ of Top-1 classification accuracy and relative loss (loss w.r.t. center) across the entire range of translational displacements on the ImageNet validation set. Accuracy measures overall performance, while Loss w.r.t. center quantifies the average performance degradation caused by translation (where a lower mean loss indicates greater robustness). The GAP-modified models, particularly Multi and Final, demonstrate significantly lower standard deviation and mean loss values compared to the baseline VGG16, confirming their superior translational robustness and consistency across the displacement grid.}
\label{tab:mean_std}
\begin{tabular*}{\textwidth}{@{\extracolsep{\fill}}lcccc}
\toprule
 & \multicolumn{2}{c}{Accuracy $\uparrow$} & \multicolumn{2}{c}{Loss w.r.t. center $\downarrow$} \\ 
\cmidrule(lr){2-3} \cmidrule(lr){4-5} 
 & Mean & Std & Mean & Std \\ 
\midrule
Base & 0.64 & 0.04 & 0.09 & 0.05 \\ 
Multi & 0.636 & 0.015 & \textbf{0.05} & \textbf{0.02} \\ 
Final & 0.664 & 0.018 & 0.06 & 0.02 \\ 
Flat & \textbf{0.67} & \textbf{0.03} & 0.09 & 0.05 \\ 
\bottomrule
\end{tabular*}
\vspace{-0.4cm}
\end{table}

\begin{center}
\begin{figure}[!h]
			\centering
			\includegraphics[width=0.8\linewidth]{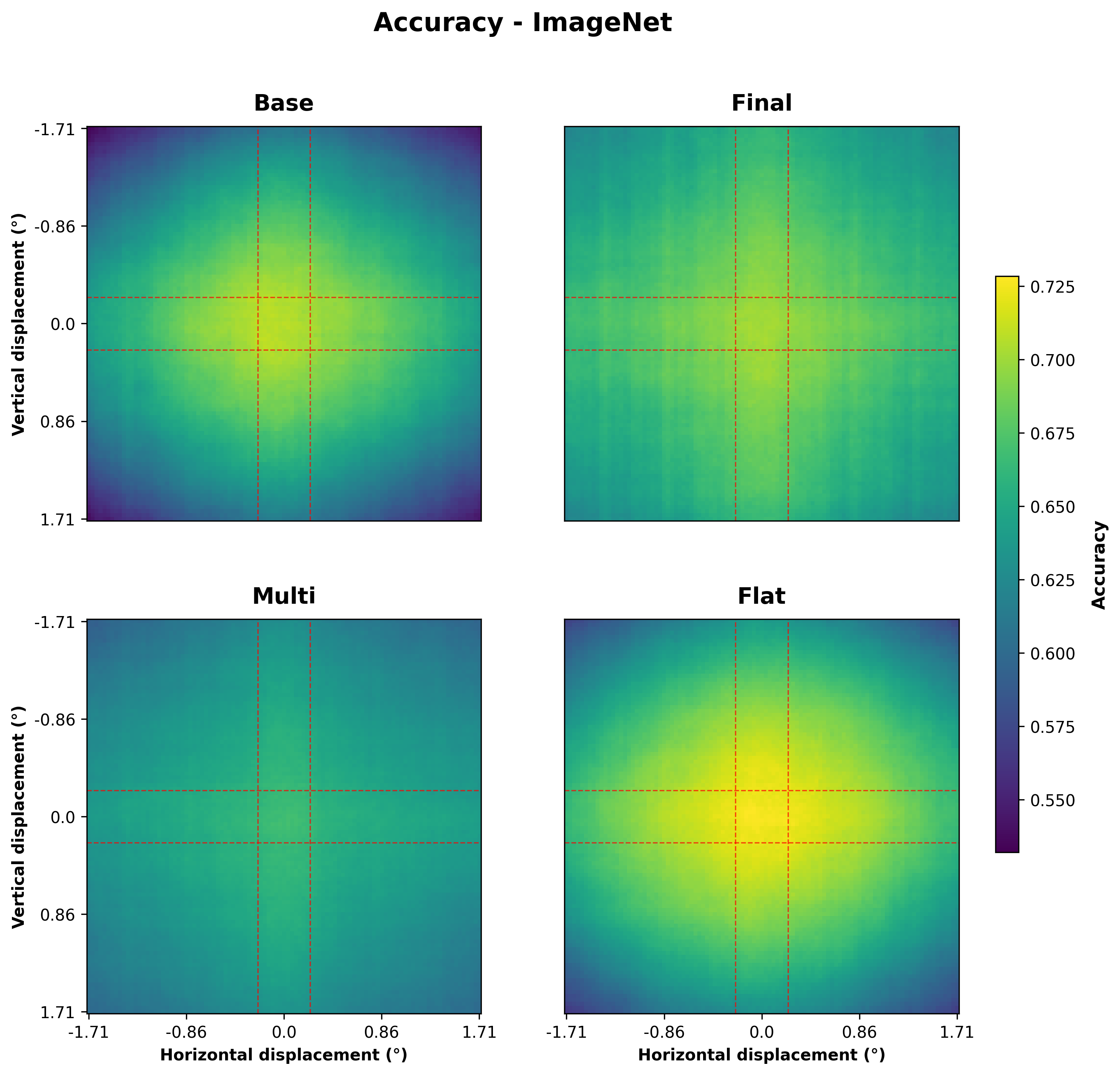}
   \caption{Heatmaps illustrating the Top-1 classification accuracy for the Base model (top left) and its three proposed translation-invariant variants (Final, Multi, Flat) across a 2D grid of horizontal and vertical displacements. The displacement is measured in degrees (or pixels/units of input space). The color scale indicates the accuracy performance, with yellow (higher values) representing better robustness. The baseline VGG16 shows a sharper drop in accuracy as displacement increases, while the proposed GAP-modified models maintain a wider, higher-accuracy core, demonstrating improved resilience and translation invariance.}
		\label{fig:Accuracy}
        %\vspace{-0.4cm}
   \end{figure}
   \end{center}

Figure~\ref{fig:Accuracy} visualizes the Top-1 Accuracy performance as a two-dimensional heatmap across the vertical and horizontal displacement space. The baseline VGG16 model exhibits a sharp peak of high accuracy concentrated around the center coordinates (zero displacement). Accuracy rapidly degrades as the displacement increases, confirming its high sensitivity to position. The Multi and Final models, however, visually show a wider and more sustained region of high accuracy, indicating that the insertion of the GAP layer prevents the abrupt performance drop away from the center.

\begin{center}
\begin{figure}[!h]
			\centering
			\includegraphics[width=0.8\linewidth]{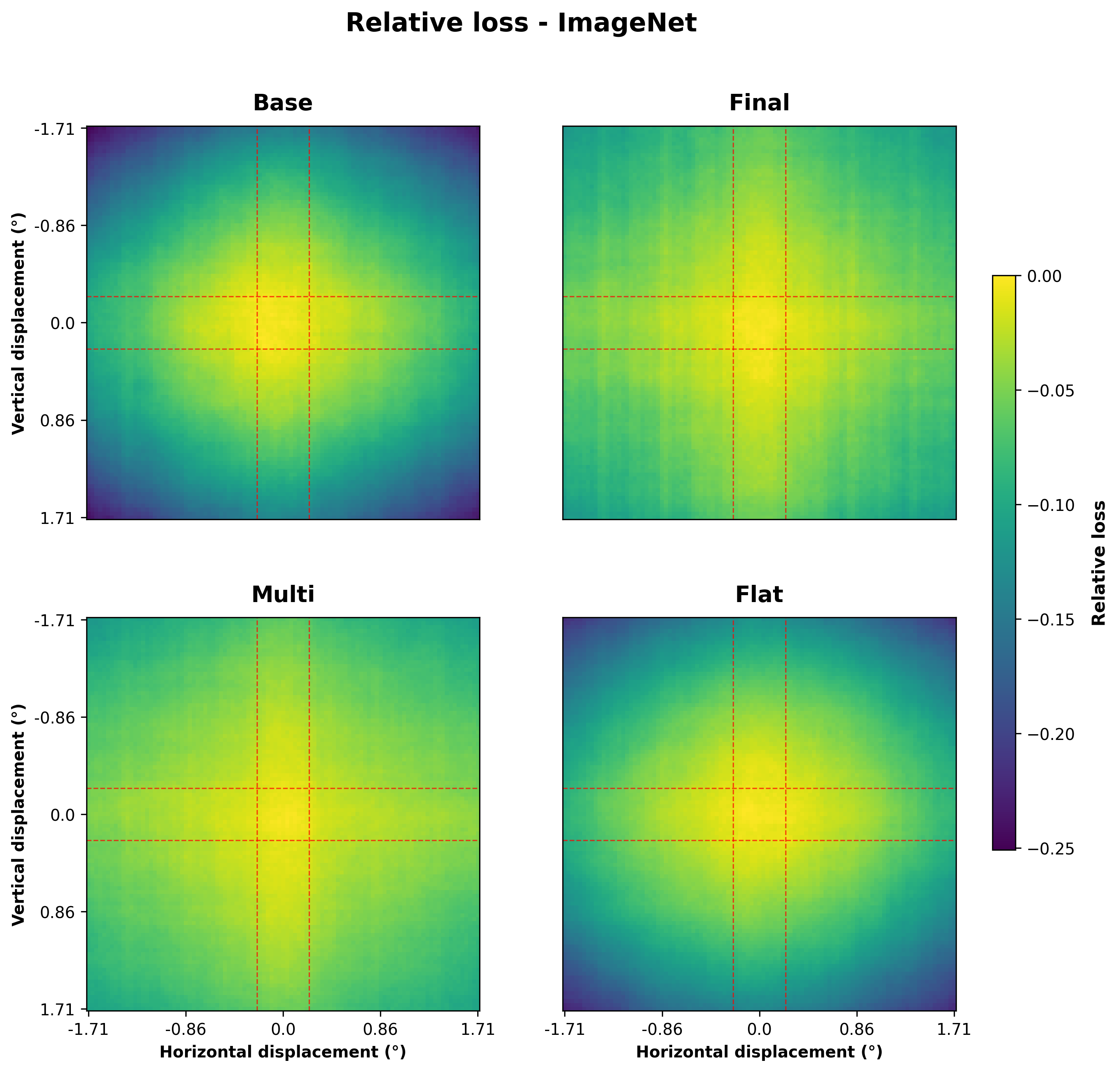}
   \caption{Heatmaps displaying the relative loss in Top-1 classification accuracy compared to the center-aligned image (zero displacement, where relative loss is 0.00) for the Base model and the three GAP-modified variants. The loss is computed as $\text{Loss}_{\text{relative}} = \text{Accuracy}_{0,0} - \text{Accuracy}_{x,y}$. The color scale indicates the magnitude of the relative loss, where yellow (closer to 0.00) signifies minimal loss and superior translation robustness, and purple indicates a greater reduction in accuracy. The baseline VGG16 (top left) shows a faster increase in relative loss away from the center (sharper purple transition). Conversely, models Multi and Flat demonstrate flatter heatmaps and maintain values closer to zero across a wider area, confirming that the GAP-based modifications are effective in achieving greater invariance to translational shifts.}
		\label{fig:Perdida}
        %\vspace{-0.4cm}
   \end{figure}
   \end{center}

For a more rigorous analysis of invariance, we examine the relative loss with respect to the center in Figure~\ref{fig:Perdida}. In this representation, a flatter profile, colored closer to zero (yellow), indicates superior robustness, as accuracy remains nearly constant despite the shift. The baseline VGG16 model displays a much steeper loss gradient, with darker colors quickly extending from the center. Conversely, the Multi and Final models exhibit the flattest loss distributions, successfully mitigating the performance deterioration caused by displacement.

In summary, both the numerical analysis in Table~\ref{tab:mean_std} and the visual evidence in Figure~\ref{fig:Perdida} confirm that the GAP-based architectural modification is highly effective. The approach not only maintains the classification capability of the VGG16 model but endows it with superior translation invariance, demonstrated by the reduced average loss and stability in performance across a wide spectrum of displacements. These results establish the Multi and Final variants as the most promising candidates for subsequent application in the perceptual domain.

\section{Results in image quality problems (IQA)}
\label{sec:results_IQA}

Following the validation of translation robustness in classification tasks, we extend the evaluation of our proposed architectures to the domain of Image Quality Assessment (IQA). By integrating the GAP-modified VGG backbones into the Learned Perceptual Image Patch Similarity (LPIPS) framework, we aim to determine if the enhanced translation invariance improves the metric's robustness without compromising its perceptual accuracy.

The evaluation of these new perceptual metrics is conducted through a dual validation strategy: (1)~First, we verify the general predictive performance of the models by benchmarking their correlation with subjective human ratings (MOS) across standard IQA datasets. This ensures that the architectural modifications preserve the metric's fundamental ability to assess image quality. (2)~Second, we perform a fine-grained analysis using the RAID dataset to compare the models' sensitivity profiles against human psychophysical thresholds. RAID has psychophysical human measures specifically with translation distortions. This step is crucial to assess whether the invariant backbones align better with the distinct "human response curves" observed under varying intensities of distortion.

To maintain consistency with the nomenclature defined in Section~\ref{sec:Mejora}, we refer to the LPIPS metrics derived from our architectures using the prefix 'L-'. Then, the standard LPIPS (based on Base) is denoted as LBase and the metric leveraging the multi-stage GAP backbone as LMulti. Note that, due to the specific feature extraction mechanism of LPIPS, the Final and Flat architectures yield equivalent metric formulations; therefore, we report results for this variant uniquely as LFlat.

\subsection{Correlation with MOS}

The performance of the proposed invariant metrics was benchmarked using the Pearson correlation across three major datasets: TID08, TID13, and KADID-10k. Table~\ref{tab:CorrMOS_app} presents these results alongside a comprehensive comparison with multiple state-of-the-art IQA metrics. As expected, the baseline LPIPS (LBaseOr) model without specific retraining shows the lowest performance among the LPIPS variants. However, once retrained on TID08, all our models demonstrate a substantial leap in correlation, with values rising from approximately 0.70 to over 0.85.

To provide a fair context, the state-of-the-art metrics included in Table~\ref{tab:CorrMOS_app} are evaluated directly, without specific retraining on TID08. It is important to highlight that our invariant architectures do not degrade the perceptual representative power of the network. In fact, both LMulti and LFlat show performance levels that are highly competitive with—and in the case of KADID-10k, superior to—the original retrained VGG-16 and rank among the top-performing solutions overall. Specifically, on the KADID-10k generalization test, the LFlat variant achieves a correlation of 0.89, significantly outperforming both the retrained baseline (0.75) and the vast majority of standard metrics.

\begin{table}[htbp]
\caption{Performance evaluation of the proposed invariant metrics and comparison with state-of-the-art IQA models. The table reports the Spearman correlation ($S$) across the TID08, TID13, and KADID-10k datasets. The top section presents the LPIPS variants: the untrained baseline (LBaseOr), the retrained baseline (LBase RT), and our invariant models (LMulti and LFlat), all of which (except LBaseOr) were trained on TID08. The bottom section displays multiple standard IQA metrics for broad context, which have not been specifically retrained on TID08. The proposed LFlat variant achieves highly competitive results overall, demonstrating superior generalization on KADID-10k.}
\label{tab:CorrMOS_app}
\begin{tabular*}{\textwidth}{@{\extracolsep{\fill}}lccc}
\toprule
Metric & TID2008 & TID2013 & KADID-10k \\ 
\midrule
LBaseOr (LPIPS) & 0.70 & 0.73 & 0.70 \\ 
LBase & 0.95 & 0.89 & 0.75 \\ 
LMulti & 0.89 & 0.87 & 0.84 \\ 
LFlat & \textbf{0.91} & \textbf{0.89} & \textbf{0.89} \\ 
\midrule
SSIMLoss & 0.66 & 0.72 & 0.65 \\ 
MultiScaleSSIMLoss & 0.74 & 0.77 & 0.68 \\ 
Information Weighted SSIMLoss & 0.81 & 0.76 & 0.72 \\ 
VIFLoss & 0.56 & 0.55 & 0.59 \\ 
FSIMLoss & 0.83 & 0.83 & 0.71 \\ 
SRSIMLoss & 0.80 & 0.77 & 0.54 \\ 
% GMSDLoss & 0.87 & 0.85 & 0.80 \\ Entrenado en TID
% MultiScaleGMSDLoss & 0.89 & 0.86 & 0.82 \\ Entrenado en TID 
VSILoss & 0.81 & 0.83 & 0.69 \\ 
DSSLoss & 0.87 & 0.84 & 0.79 \\ 
%HaarPSILoss & 0.90 & 0.89 & 0.84 \\ Entrenado en TID
MDSILoss & 0.82 & 0.84 & 0.83 \\ 
PieAPP & 0.41 & 0.50 & 0.24 \\ 
DISTS & 0.80 & 0.83 & 0.86 \\ 
StyleLoss & 0.23 & 0.27 & 0.38 \\ 
Content Loss & 0.67 & 0.70 & 0.65 \\ 
\bottomrule
\end{tabular*}
\vspace{-0.4cm}
\end{table}

\subsection{Correlation with human psychophysics}
\label{sec:psychophysics}

While Mean Opinion Scores (MOS) provide a scalar value for image quality, they do not fully capture the suprathreshold behavior of the human visual system—specifically, how perceived quality degrades as distortion intensity increases. To validate whether our proposed invariant models replicate this psychophysical behavior, we utilized the RAID dataset \cite{RAID}, which provides human response curves to affine image transformations. The human response curves were derived using Maximum Likelihood Difference Scaling (MLDS).

To evaluate the models, we adopt a methodology that focuses on the integration of local contrast changes. Instead of a direct comparison against the pristine reference image, we measure the local distance between images at consecutive translation degrees ($D(I_{n-1}, I_{n})$). The global response curve is then constructed by computing the cumulative sum of these local, step-wise distances. This approach is specifically designed to capture the monotonic growth and cumulative nature of human perception as the intensity of the geometric distortion grows. A comprehensive comparison with other standard evaluation methodologies is provided in Appendix B.

\begin{figure}[!h]
			\centering
			\includegraphics[width=1\linewidth]{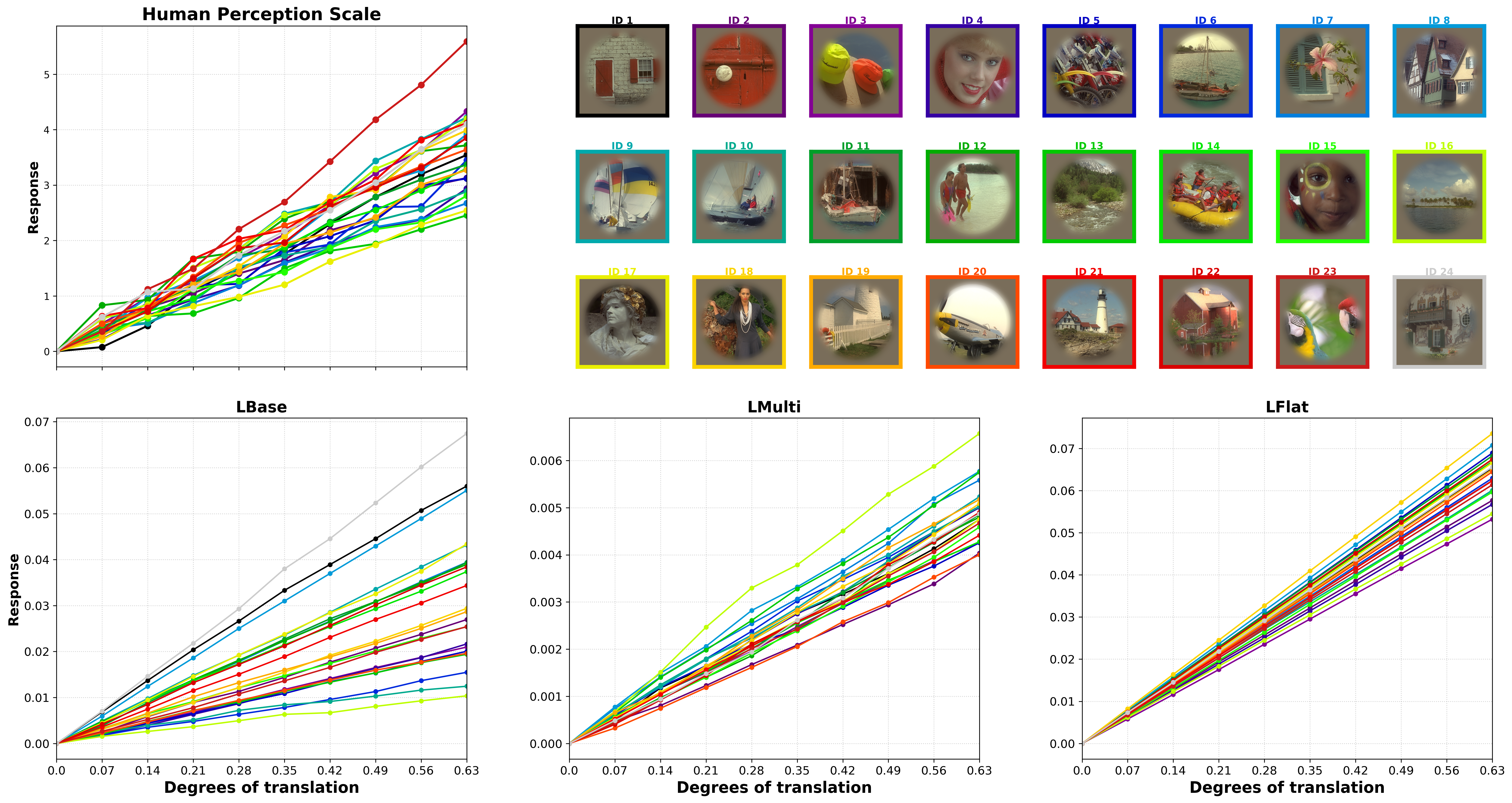}
   \caption{Comparison of Human Response Curves vs. Model Response Curves under Translation. (Top row) Human perception curves from the RAID dataset and visual legend. (Rows 2-4) Response curves generated by the LBase, LMulti, and LFlat models. The plots illustrate the proposed methodology based on the integration of local contrast changes, showing a strong alignment with human monotonic growth, particularly for the LFlat variant. Appendix~\ref{app:apndB} provides the full comparison across all tested methodologies.}
		\label{fig:LPIPSMOD}
        \vspace{-0.4cm}
   \end{figure}

The results for this evaluation scheme are detailed in Table~\ref{tab:Numeritis}. As visualized in Figure~\ref{fig:LPIPSMOD}, this method produces response curves that strikingly resemble the human profile. Specifically, the L-Flat variant achieves exceptional performance ($S=0.95, P=0.93$), slightly outperforming the L-Base model and the L-Multi variant.

The quantitative analysis in Table~\ref{tab:Numeritis} reinforces this alignment, where L-Flat shows the minimal mean error ($\mu=0.027$) and the tightest variance ($\sigma=0.009$) among all architectures. These results confirm that enforcing translation invariance via L-Flat does not distort the model's psychophysical alignment; on the contrary, it provides a more biologically plausible numerical replication of human suprathreshold response.

\begin{table}[htbp]
\caption{Quantitative differences and correlation coefficients between the model predictions and the human psychophysical curves from the RAID dataset. The table reports the mean ($\mu$) and standard deviation ($\sigma$) of the point-by-point absolute differences (after [0, 1] normalization), alongside the Spearman ($S$) and Pearson ($P$) correlations.}
\label{tab:Numeritis}
\begin{tabular*}{\textwidth}{@{\extracolsep{\fill}}lccc}
\toprule
 & LBase & LMulti & LFlat \\ 
\midrule
\multicolumn{4}{l}{\textit{Quantitative differences}} \\
$\quad \mu$ & 0.027 & 0.033 & \textbf{0.027} \\ 
$\quad \sigma$ & 0.010 & 0.013 & \textbf{0.009} \\ 
\midrule
\multicolumn{4}{l}{\textit{Correlation coefficients}} \\
$\quad S$ & 0.83 & 0.93 & \textbf{0.95} \\ 
$\quad P$ & 0.75 & 0.91 & \textbf{0.93} \\ 
\bottomrule
\end{tabular*}
\vspace{-0.4cm}
\end{table}

\section{Discussion}
\label{sec:Discussion}

The results presented in this study offer a comprehensive validation of the proposed "Online Architecture" approach, demonstrating that structural modifications using Global Average Pooling (GAP) can effectively induce translation invariance in Convolutional Neural Networks without the need for extensive data augmentation or retraining from scratch. This is visually confirmed by the flat accuracy heatmaps in Figure~\ref{fig:Accuracy} and quantitatively supported by the drastic reduction in relative loss. However, the residual vertical banding observed in the high-resolution analysis (Section~\ref{sec: results_MNIST}) highlights an intrinsic limitation: the discrete sampling of pooling layers introduces aliasing that even GAP cannot fully eliminate. This suggests that, while GAP solves the macroscopic translation problem, perfect invariance at the pixel level requires addressing the sampling theorem violations in downsampling layers.\\
A common concern in robust optimization is the potential loss of discriminative power or increased computational cost. Our results refute this for the proposed method. The Final variant not only improved robustness but did so while reducing the trainable parameter count by over 98\% (Table~\ref{tab:Parametros}). This implies that the massive fully connected layers in standard VGG architectures are largely redundant for classification tasks that do not require explicit localization, and their removal actually benefits model stability.\\
Standard metrics like LPIPS often overestimate the perceptual difference caused by imperceptible shifts. Our results with the LFlat variant on the TID and KADID datasets demonstrate that it is possible to build a metric that is mathematically robust to translation while maintaining—and even exceeding—the correlation with human judgment ($S=0.89$ on KADID). Furthermore, the psychophysical validation using the RAID dataset (Table~\ref{tab:Numeritis}) confirms that this architectural invariance aligns with the human visual system's own indifference to small shifts, providing a more biologically plausible model of image quality.

\section{Conclusion}
\label{sec:conclusion}
This work addresses the fundamental lack of translation invariance in standard Convolutional Neural Networks, identifying the fully connected layers as the primary source of positional sensitivity. We proposed and evaluated a series of architectural modifications based on the strategic insertion of Global Average Pooling (GAP) layers into the VGG-16 backbone at specific points to tackle this issue.\\
The proposed GAP-modified models achieve superior translation invariance compared to the baseline. Specifically, the Multi and Final variants maintain high classification accuracy across the entire range of translational displacements, effectively mitigating the performance drop observed in standard models.\\
We demonstrated that the GAP layer acts as a spatial regularizer that decouples semantic content from spatial position. Importantly, this robustness is achieved with a drastic reduction in model complexity, decreasing the amount of trainable parameters by over 98\% compared to the baseline, without compromising classification capability. However, we also identified that discrete pooling operations introduce a residual, periodic sensitivity (aliasing) that remains a limiting factor.\\
When applied to the LPIPS framework, the LFlat architecture proved to be the most effective, generalizing better than the original model on diverse IQA datasets (KADID) and accurately replicating human suprathreshold response curves (RAID).\\
In summary, this study confirms that enforcing translation invariance through architectural constraints is a superior strategy to data augmentation alone. The proposed modifications provide a lightweight, robust, and perceptually accurate backbone that is ready for deployment in real-world classification and image quality assessment applications.

% --- SECCIÓN DE DECLARACIONES OBLIGATORIA PARA SPRINGER ---
\vspace{-0.2cm}
\section{Declarations}
\vspace{-0.2cm}
\textbf{Funding:} Supported by MICIIN/FEDER/UE (Grants PID2020-118071GB-100, PDC2021-121522-C21) and Generalitat Valenciana (GV/2021/074, CIPROM/2021/056, CIAPOT/2021/9, CIACIF/2023/223). Computer resources provided by Artemisa (EU ERDF, Comunitat Valenciana) with technical support from IFIC (CSIC-UV). \textbf{Competing Interests:} The authors declare that they have no conflict of interest. \textbf{Data and Code Availability:} The datasets (TID08, TID13, KADID-10k, RAID) and the code to reproduce the results are publicly available at \url{https://github.com/Rietta5/Translation_Invariant_CNNs}. \textbf{Author Contributions:} N.A.-B.: Conceptualization, Data Curation, Formulation, Investigation, Methodology, Software, Visualization, Writing - Original draft. J.V.-T.: Investigation, Software, Writing - review \& editing. P.D.-O.: Investigation, Software, Visualization, Writing - review \& editing. V.L. \& J.M.: Conceptualization, Formulation, Methodology, Supervision, Writing - Original draft.

\begin{appendices}
\section{Results for other transformations} %related
\label{app:apndB}

While the primary focus of this work is Translation Invariance, for completeness, we evaluate the proposed architectures against other common affine perturbations: Rotation and Scaling. Since the Global Average Pooling (GAP) mechanism is mathematically designed to address spatial shifts but does not inherently provide invariance to rotation or scale changes, we do not expect the proposed modifications to outperform the baseline in these domains.
\vspace{-0.4cm}
\begin{center}
\begin{figure}[!h]
\centering
			\includegraphics[width=0.78\linewidth]{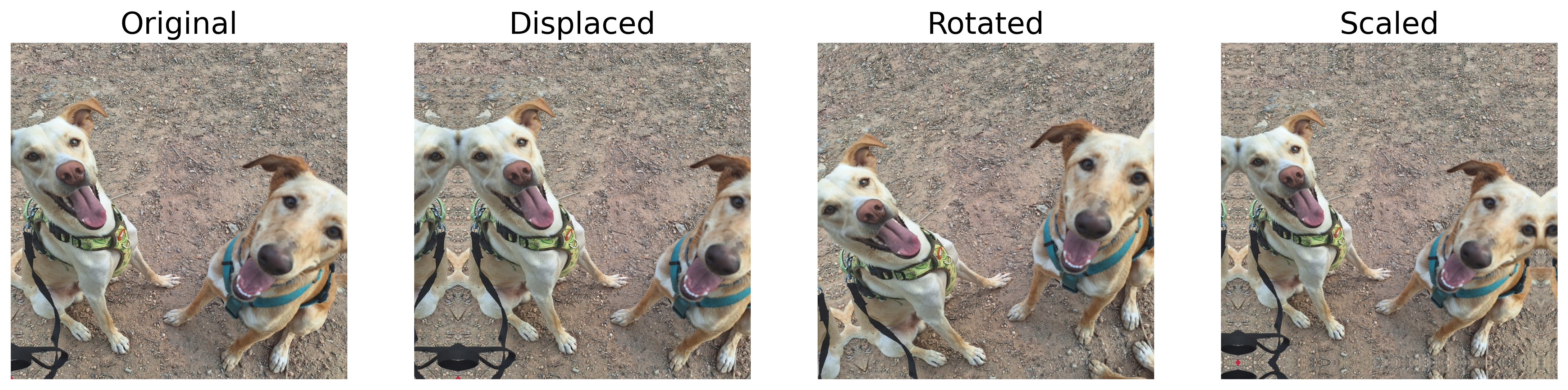}	
		
		\caption{Visual examples of the affine transformations applied in the control experiments. From left to right: the original image, translation (primary focus), rotation ($20^{\circ}$), and scaling (factor of 0.9)}
		\label{fig:distorsiones_TA}
        %\vspace{-0.4cm}
	\end{figure}
\end{center}

\vspace{-1cm}

\begin{center}
\begin{figure}[!h]
			\centering
			\includegraphics[trim=0 5 0 5, clip,width=1\linewidth]{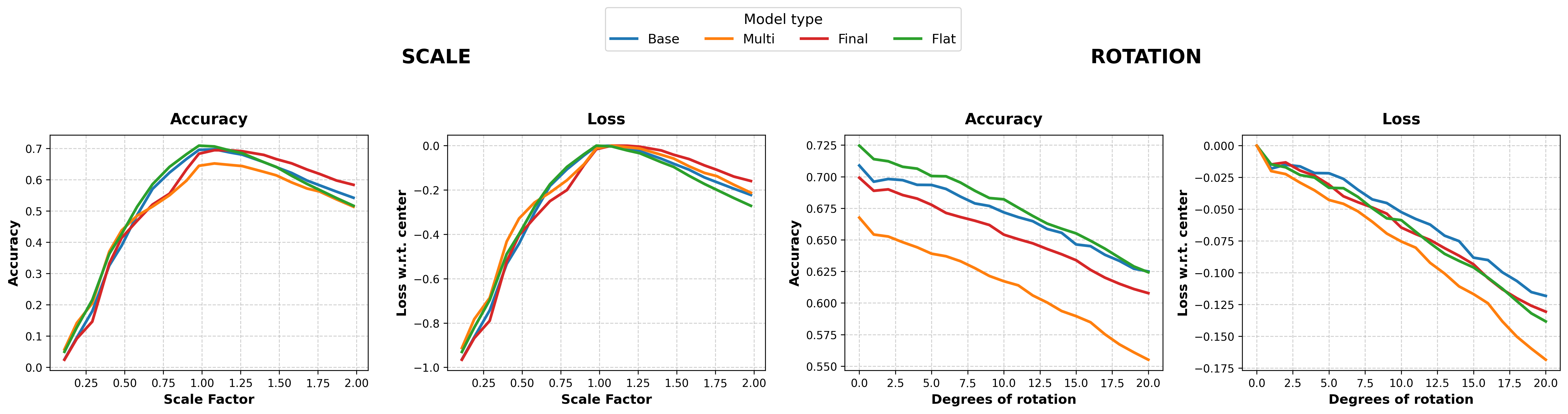}
   \caption{Performance comparison of the Base model and its variants (Multi, Final, Flat) under additional affine transformations: Scaling (top row) and Rotation (bottom row). The left column plots show Top-1 Accuracy versus the transformation magnitude, while the right column shows the Loss relative to the non-transformed center. Unlike the translation experiments, the GAP-based variants do not outperform the baseline here. In fact, under rotation, the Base and Flat models exhibit superior stability compared to Multi, suggesting that preserving spatial feature layout is beneficial for orientation robustness. Under scaling, all models display a similar bell-shaped degradation, confirming that the invariance gains reported in this work are specific to translation.}
		\label{fig:Rotacion_Escala}
        %\vspace{-0.4cm}
   \end{figure}
   \end{center}

Figure~\ref{fig:Rotacion_Escala} presents the accuracy and loss landscape for both transformations. \textbf{Rotation}: As observed in the bottom row of Figure B.1, all models suffer a linear degradation in performance as the rotation angle increases from $0^{\circ}$ to $20^{\circ}$.

\begin{itemize}
    \item The Base and Flat models exhibit the highest robustness, maintaining an accuracy above 0.62 at $20^{\circ}$. This suggests that preserving the spatial layout of features (via the Flatten layer or the original dense layers) is beneficial for rotational stability.
    \item The Multi variant shows the lowest performance, starting with a lower initial accuracy and degrading at a similar rate.
    \item The Final model sits in between.This confirms that the GAP layers, while solving translation, do not confer rotational invariance, as standard convolution filters themselves are orientation-selective.
\end{itemize}

\textbf{Scaling}:The top row of Figure B.1 shows the response to scaling factors ranging from $0.1\times$ to $2.0\times$.
\begin{itemize}
    \item All architectures display a characteristic "bell-shaped" performance curve centered at the original scale ($1.0\times$).
    \item The Base and Flat models show slightly better resilience to downscaling (factors $< 1.0$) compared to the GAP-heavy variants.
    \item The Multi model again exhibits a slightly lower overall accuracy envelope across the scale range.
\end{itemize}

\vspace{-0.4cm}

\vspace{-1cm}

   In conclusion, these results demonstrate the specificity of the GAP modification: the architectural enhancement is highly effective in increasing robustness against translation, but it does not significantly improve robustness against other affine transformations like rotation and scaling. This underscores that the benefit is directly tied to the spatial averaging mechanism of GAP and its ability to decouple feature presence from spatial location, a mechanism that does not inherently address the feature distortion caused by rotation or the resolution changes caused by scaling.

\section{Comparison of Psychophysical Evaluation Methodologies}
\label{app:apndB}

To ensure the robustness of our perceptual validation, we compared the methodology presented in Section \ref{sec:psychophysics} (hereafter referred to as Sequential) with three alternative schemes:
\begin{itemize}
    \item Original - Distorted: Direct metric distance between the reference and the distorted image.
    \item Cumulative Sum: Accumulation of distances measured always against the reference.
    \item MLDS Simulation: An adaptation of the Maximum Likelihood Difference Scaling algorithm with fixed internal noise ($\sigma \approx 0.29$).
\end{itemize}

\begin{figure}[!h]
			\centering
			\includegraphics[width=0.95\linewidth]{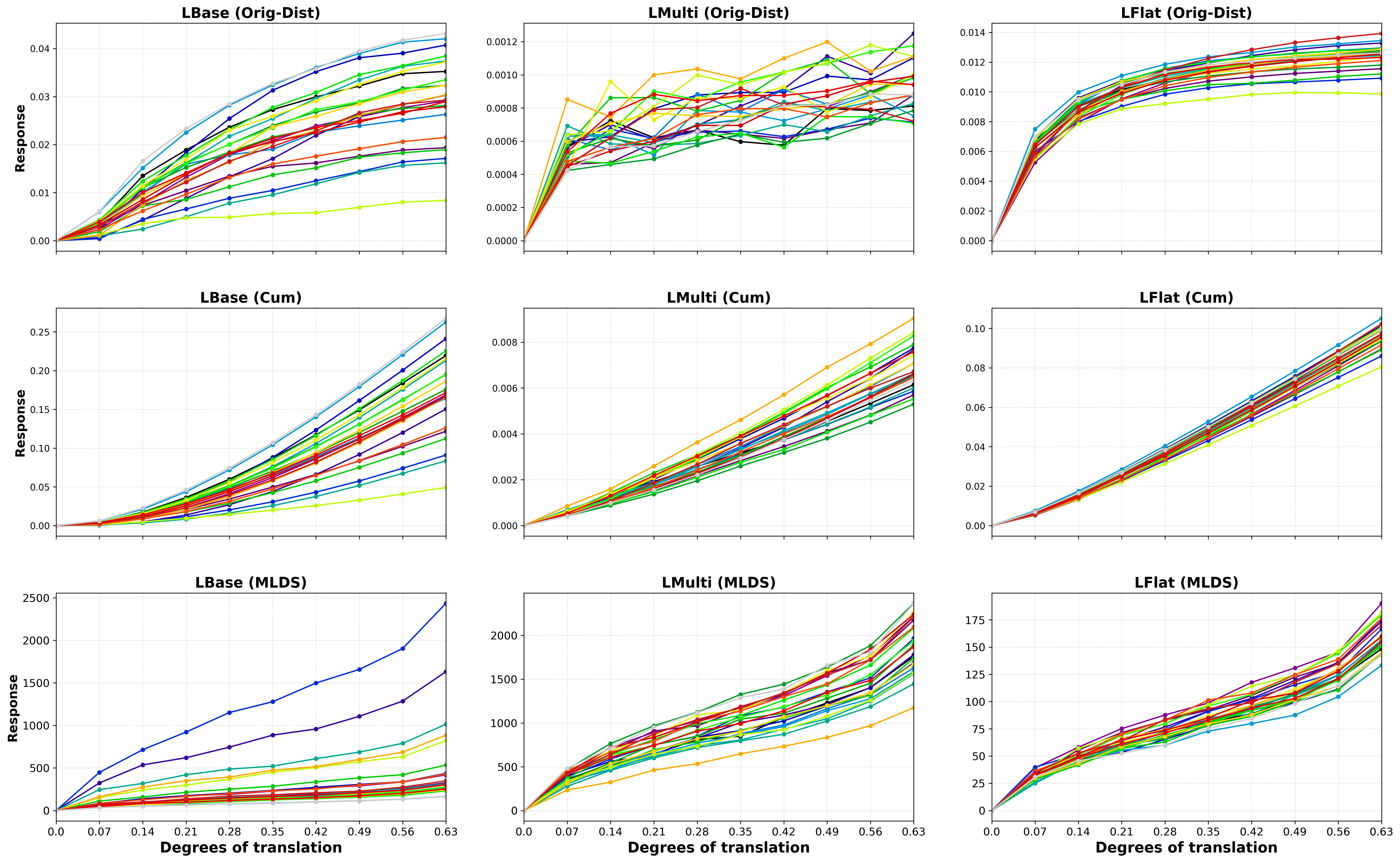}
   \caption{Comparison of Human Response Curves vs. Model Response Curves under Translation. Response curves generated by the LBase, LMulti, and LFlat models using four distinct calculation methodologies: (1) Standard Distance (direct comparison between original and distorted images), (2) Cumulative Sum of standard distances, (3) MLDS Simulation (mimicking the Maximum Likelihood Difference Scaling algorithm with a fixed internal noise $\sigma=0.29$, explicitly calibrated using the first-order distortion response). The x-axis represents the translation magnitude in degrees, increasing by $0.07^\circ$ at each step. The calibrated MLDS and Cumulative Sequential approaches generate curves that are strongly aligned with the monotonic growth observed in human perception, particularly for the LFlat variant.}
		\label{fig:LPIPSMOD_All}
        %\vspace{-0.4cm}
   \end{figure}

%\looseness=-1
As shown in the extended results Figure~\ref{fig:LPIPSMOD_All}, Tables~\ref{tab:DifCurve} and ~\ref{tab:CorrValor}, while methods like MLDS also yield high correlations, the Sequential approach described in the main text proves to be the most accurate in terms of numerical distance ($\mu$) and curve-shape reproduction.

\begin{table}[htbp]
\caption{Quantitative differences between the model response curves and the human psychophysical curves from the RAID dataset. To ensure scale equivalence, all curves were first normalized to a [0, 1] range by dividing by their respective maximum values. The table reports the mean ($\mu$) and standard deviation ($\sigma$) of the point-by-point absolute differences across all images. Lower mean values indicate a tighter numerical alignment and a more accurate shape reproduction of human perception.}
\label{tab:DifCurve}
\begin{tabular*}{\textwidth}{@{\extracolsep{\fill}}lccc}
\toprule
 & LBase & LMulti & LFlat \\ 
\midrule
\multicolumn{4}{l}{\textit{Orig - Dist}} \\
$\quad \mu$ & 0.09 & 0.23 & 0.28 \\ 
$\quad \sigma$ & 0.03 & 0.05 & 0.02 \\ 
\midrule
\multicolumn{4}{l}{\textit{Cumsum}} \\
$\quad \mu$ & 0.11 & 0.045 & 0.043 \\ 
$\quad \sigma$ & 0.02 & 0.019 & 0.017 \\ 
\midrule
\multicolumn{4}{l}{\textit{MLDS}} \\
$\quad \mu$ & 0.058 & 0.056 & 0.057 \\ 
$\quad \sigma$ & 0.012 & 0.011 & 0.011 \\ 
\midrule
\multicolumn{4}{l}{\textit{Sequential}} \\
$\quad \mu$ & \textbf{0.027} & \textbf{0.033} & \textbf{0.027} \\ 
$\quad \sigma$ & \textbf{0.010} & \textbf{0.013} & \textbf{0.009} \\ 
\bottomrule
\end{tabular*}
\end{table}

\begin{table}[htbp]
\caption{Spearman (S) and Pearson (P) correlation coefficients between the human response curves and the curves in RAID by the three models (Original, GAP, LFlat) using the four calculation methodologies defined in Figure~\ref{fig:LPIPSMOD}. The MLDS method yields the highest correlations across all models, confirming it as the most accurate proxy for comparing deterministic metrics with human psychophysics. Notably, the LFlat variant achieves the highest overall correlation under the MLDS method, demonstrating that our modifications preserve the model's ability to replicate human suprathreshold response.}
\label{tab:CorrValor}
\begin{tabular*}{\textwidth}{@{\extracolsep{\fill}}lccc}
\toprule
 & LBase & LMulti & LFlat \\ 
\midrule
\multicolumn{4}{l}{\textit{Orig - Dist}} \\
$\quad S$ & 0.82 & 0.72 & 0.89 \\ 
$\quad P$ & 0.79 & 0.69 & 0.78 \\ 
\midrule
\multicolumn{4}{l}{\textit{Cumsum}} \\
$\quad S$ & \textbf{0.92} & 0.93 & \textbf{0.95} \\ 
$\quad P$ & \textbf{0.84} & 0.90 & \textbf{0.94} \\ 
\midrule
\multicolumn{4}{l}{\textit{MLDS}} \\
$\quad S$ & 0.65 & 0.92 & 0.95 \\ 
$\quad P$ & 0.34 & 0.90 & 0.92 \\ 
\midrule
\multicolumn{4}{l}{\textit{Sequential}} \\
$\quad S$ & 0.83 & \textbf{0.93} & 0.95 \\ 
$\quad P$ & 0.75 & \textbf{0.91} & 0.93 \\ 
\bottomrule
\end{tabular*}

\end{table}
\end{appendices}

\clearpage

\begingroup
\small
\setlength{\bibsep}{0.1pt} %
\bibliography{sn-bibliography}% common bib file
\endgroup

\end{document}